# Auditable by Construction: An Ontology-Driven Framework for Trustworthy LLM Analytics in Enterprise Finance

Sergiy Lunyakin, Senior Member IEEE

Independent Researcher — Seattle, USA

slunyakin@outlook.com

## Abstract

Enterprise adoption of large language models (LLMs) in finance is constrained less by fluency than by trust: in Financial Planning and Analysis (FP&A) and other regulated workflows, an answer is usable only if it is traceable to authoritative sources, explainable in domain terms, and auditable after the fact. This paper argues that retrieval-augmented generation (RAG) for enterprise finance should be evaluated on auditability alongside accuracy, and presents a framework designed to make grounded responses auditable by construction. The Knowledge-Driven Analytics Framework (KDAF) builds ontology-driven knowledge systems through six iterative stages — problem-centric scoping via competency questions, ontology bootstrapping through a minimum viable graph, schema-guided knowledge extraction, contextual knowledge representation with typed relevance and provenance annotations, hybrid human-in-the-loop validation, and Context-Aware Relevance Propagation (CARP) for graph-based evidence retrieval — so that every retrieved fact carries its relationship type, confidence, and source lineage.

A proof-of-concept evaluation on FinanceBench (145 questions) compares KDAF against zero-context inference, BM25 sparse retrieval, concept-weighted lexical retrieval over text and tables, and ungrounded graph traversal. Three findings structure the results. First, retrieval is necessary: zero-context inference reaches 4.1% correctness against 10-12% for every retrieval-augmented condition. Second, on answer correctness the retrieval conditions are statistically indistinguishable — KDAF and BM25 differ by -0.007 (95% CI [-0.021, 0.000]) — so accuracy alone does not justify the cost of structured retrieval on single-document filings questions, a negative result we report explicitly. Third, on the properties auditability depends upon, the ordering reverses: KDAF attains the highest citation traceability F1 (0.515), exceeding ungrounded graph traversal by +0.027 (95% CI [0.006, 0.050]) and BM25 by +0.052 (95% CI [0.024, 0.083]) — intervals that exclude zero. Separately, graph-structured retrieval of either kind admits no evidence from outside the question subject entity (0 of 426 and 0 of 424 items, against 16.8% and 20.2% for the lexical baselines), because the entity model supplying the provenance chain also makes that boundary expressible as an enforced constraint; every selected item resolves to a complete provenance chain. We argue that auditability, not raw question-answering accuracy, is the axis on which ontology-grounded retrieval earns its construction cost, and outline the FP&A-aligned evaluation this motivates.

## 1. Introduction

### 1.1 Motivation

Analytical models used in financial decision-making operate under explicit supervisory expectations. Longstanding regulatory guidance on model risk management requires that models be documented in sufficient detail for an independent party to understand their operation, that their inputs and outputs be traceable, and that they be subject to effective challenge — critical analysis by objective, informed reviewers able to identify limitations and compel change [Board of Governors of the Federal Reserve System & OCC, 2011]. Guidance specific to artificial intelligence extends the same logic: accountability, transparency, explainability, and interpretability are named characteristics of trustworthy AI systems, and the framework treats the ability to explain a system's output as a precondition for managing its risk rather than as an optional enhancement [NIST, 2023]. A system that produces financial analysis without a reviewable account of the evidence behind it does not satisfy these expectations, however fluent its output.

This requirement is not merely procedural. Research on human reliance upon automated systems shows that appropriate trust depends on the operator's ability to calibrate confidence against the system's actual competence, which in turn depends on the system exposing information sufficient to support that calibration; systems that conceal their basis produce either misplaced reliance or wholesale rejection [Lee & See, 2004]. Work on interpretability in machine learning makes a parallel argument, distinguishing explanations that satisfy a real evaluative need from those that merely appear satisfying [Doshi-Velez & Kim, 2017]. In an audit-bearing setting, the evaluative need is specific: a reviewer must be able to determine which evidence supported a conclusion and why that evidence was selected.

Against these expectations, practitioner evidence indicates that current FP&A capability falls short. Industry surveys report that substantial portions of FP&A effort are consumed by data collection and validation rather than analysis, that only a small fraction of finance organizations have achieved meaningful workflow automation, and that demand for data-centric finance skills is accelerating faster than available supply [FP&A Trends, 2025; McKinsey, 2024; Deloitte, 2025; FERF Research, 2024]. These are practitioner surveys rather than peer-reviewed studies, and their sampling frames and response rates vary; they are cited here as indicative of practice rather than as precise measurements. Generative AI offers one avenue of relief, but adoption inside finance functions remains largely in early or pilot phases [Agrawal et al., 2024].

The root challenge is therefore not access to language models but the absence of a principled method for grounding their outputs such that every generated claim is auditable: traceable to its sources, typed by its relationship to the question, and reviewable by a human controller. A variance explanation that cannot be traced to authoritative sources is not merely unhelpful in a finance function — it is unusable, because FP&A conclusions feed decisions carrying fiduciary, regulatory, and audit consequences. This paper develops and evaluates such a method.

## 1.2 Scientific Context and Gap

Knowledge graphs (KGs) have established themselves as a foundation for enterprise information systems, enabling machine reasoning over heterogeneous, multi-source data [Galkin et al., 2017; Meckler, 2024]. Practical methodologies for constructing KGs from structured and semi-structured enterprise data have been documented [Li et al., 2023], and their role in grounding LLM inference has attracted growing research attention, most notably through graph-based retrieval-augmented generation (GraphRAG) [Edge et al., 2024; Han et al., 2025].

In financial applications, KG use has concentrated primarily in fraud detection, credit risk assessment, anti-money laundering, and regulatory compliance [Fang et al., 2025]. Recent work has begun applying schema-guided, expert-defined extraction to authoritative financial documents such as SEC filings [Abhinav et al., 2025]. The integration of knowledge graphs into FP&A workflows — specifically for analytical reasoning over planning cycles, variance analysis, and narrative explanation — remains largely unexplored in the research literature.

LLMs have demonstrated strong performance in natural language to structured query translation [Mohammadjafari et al., 2024], and enterprise-grade text-to-SQL systems have addressed concerns of accuracy and latency at scale [Kumar et al., 2025]. However, these approaches are predominantly schema-driven and lack embedded semantic context, reducing their ability to produce explanations that are consistent, traceable, and meaningful to financial decision-makers [Franken, 2025]. Retrieval-augmented generation (RAG) mitigates some limitations of parametric LLM knowledge [Lewis et al., 2020; Gao et al., 2023], but standard vector-based RAG lacks the structured relational reasoning required to surface causal chains and multi-hop relationships inherent in financial planning data [Gao et al., 2023].

Equally important, and less examined, is what a retrieval architecture can expose about its own behavior. Flat retrievers — sparse or dense — justify their selections with a similarity score, which is not an explanation a financial controller or auditor can act upon. Ontology engineering provides a complementary path: by explicitly modeling domain concepts, relationships, and provenance, ontologies enable structured reasoning, and structured reasoning leaves an inspectable trace [Kommineni et al., 2024]. To the best of the author's knowledge, no existing work fully addresses the combination of: (a) ontology-driven knowledge graph construction aligned with FP&A competency questions, (b) context-aware graph retrieval for LLM grounding, and (c) systematic evaluation of this approach — on auditability as well as accuracy — against ungrounded retrieval baselines in a financial reasoning context.

### 1.3 Research Questions

**RQ1:** Can a problem-centric, ontology-driven methodology reliably guide the construction of semantically coherent knowledge graphs from heterogeneous enterprise financial data?

**RQ2:** Does ontology-grounded graph retrieval produce more contextually complete and traceable responses to FP&A analytical questions than ungrounded retrieval baselines?

**RQ3:** To what extent does ontology-grounded retrieval expose an auditable reasoning path — seed entities, typed traversal, and source provenance — that flat retrieval architectures structurally cannot provide?

### 1.4 Contributions

- We propose KDAF, a six-stage methodology for constructing ontology-driven knowledge graphs aligned with FP&A analytical requirements, addressing the cold-start problem through a minimum viable graph approach.
- We introduce Context-Aware Relevance Propagation (CARP), a graph-based evidence retrieval algorithm that combines ontology-defined relationship weights with dynamic relevance thresholding to optimize retrieved context for LLM grounding, and that emits a complete retrieval trace — seed entities with match reasons, typed traversal paths, threshold decisions, and end-to-end source provenance — as a first-class output.
- We present a proof-of-concept evaluation on FinanceBench (145 questions) against four baselines, reporting a negative and a positive result with equal prominence. Negative: sparse lexical retrieval matches graph-structured retrieval on answer correctness for single-document filings questions, so accuracy alone does not justify the construction cost. Positive: ontology-grounded retrieval attains significantly higher citation traceability than both ungrounded graph traversal and BM25; and graph-structured retrieval, ontology-grounded or not, admits no evidence from outside the question subject entity and resolves every selected item to a complete provenance chain. Together these argue for auditability, rather than accuracy, as the axis on which structured retrieval earns its cost in this setting.

## 2. The Knowledge-Driven Analytics Framework (KDAF)

The proposed framework provides a systematic, repeatable process for constructing ontology-driven knowledge systems that ground LLMs in verified, context-rich financial knowledge. It addresses the cold-start problem — the challenge of building a meaningful knowledge base from heterogeneous enterprise data without an existing schema — through a structured, business-aligned methodology. The framework consists of six sequential yet iterative stages, enhanced by specialized algorithms and validation techniques.

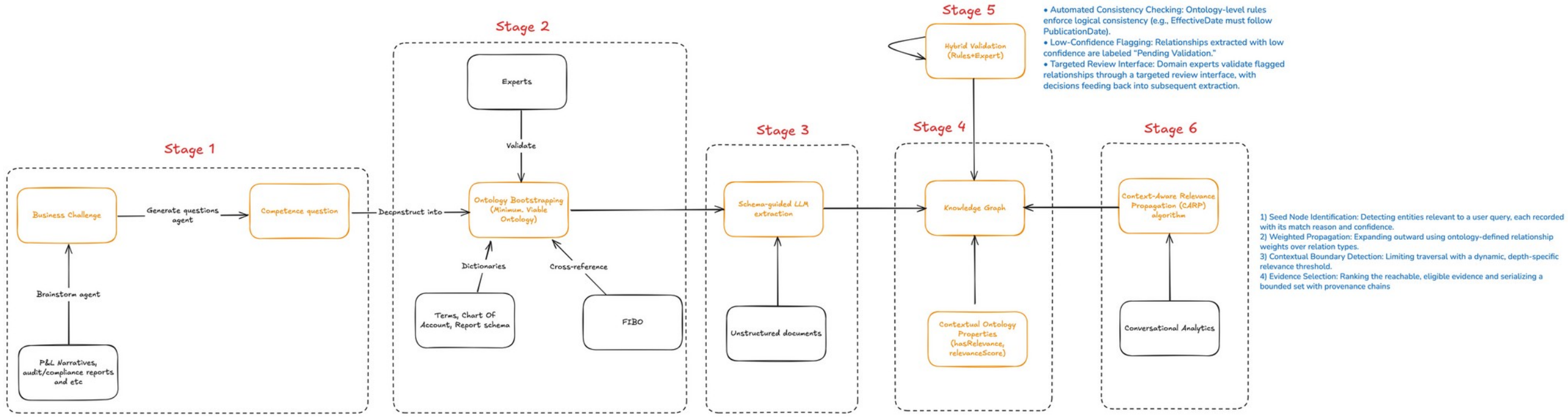

*Figure 1. The six-stage Knowledge-Driven Analytics Framework, from business challenge to context-aware retrieval. Stages 3, 4 and 6 are instantiated in the evaluation reported in Section 4; Stages 1, 2 and 5 presuppose domain experts and an organizational context that a public benchmark does not supply and are not exercised here (Section 4.1). The CARP steps listed for Stage 6 correspond to Algorithm 1.*

## 2.1 Stage 1: Problem-Centric Scoping

The process begins with a clear articulation of a business problem rather than a data problem. Following the problem-centric scoping principle, finance professionals, analysts, and data engineers collaboratively define key challenges — for example, "Why did operational expenses increase in Q2?" or "What are the drivers behind declining margin contribution in product line A?" Each challenge is transformed into a structured set of Competency Questions, formally defining the scope of the ontology and the knowledge graph [Kommineni et al., 2024]. This step ensures every subsequent model component directly aligns with business value, preventing the creation of disconnected or purely technical schemas.

## 2.2 Stage 2: Ontology Bootstrapping

To overcome the cold-start barrier, competency questions are deconstructed into their fundamental semantic components — entities, metrics, and relationships — forming a Minimum Viable Graph (MVG). For example, the question "What are the primary cost drivers of Q2 OPEX?" leads to the identification of entities such as ExpenseCategory, CostCenter, and Period, and relationships such as impacts, aggregates, and belongsTo. The resulting MVG becomes the scaffold for the knowledge graph and is validated with subject matter experts. To enhance semantic rigor, the MVG is cross-referenced with existing upper ontologies such as the Financial Industry Business Ontology (FIBO) [EDM Council, 2024], ensuring consistency and reuse across financial domains.

## 2.3 Stage 3: Schema-Guided Knowledge Extraction

With the MVG in place, the framework employs schema-guided LLM extraction to populate the graph from heterogeneous data sources, including financial statements, internal reports, and regulatory filings. The LLM operates under schema constraints, mapping textual content to predefined ontology classes (e.g., Revenue, Expense, BusinessEvent) rather than performing free-form entity extraction. This guided extraction approach improves precision and relevance by anchoring LLM outputs in the predefined semantic structure. New entities discovered during extraction can trigger iterative ontology evolution, ensuring adaptability to new financial concepts or data sources.

## 2.4 Stage 4: Contextual Knowledge Representation

The framework extends standard OWL/RDF representation [W3C, 2012; W3C, 2014] by introducing contextual relevance and provenance annotations. For example, a relationship between NewProductLaunch and Q3RevenueReport can be annotated with a contextual relevance type (e.g., myonto:hasContextualRelevance myonto:CausalDriver), a confidence score, and a source document reference. These constructs allow the framework to express why a fact matters — whether causal, supporting, or correlational — rather than merely stating it. Provenance annotations are modeled as first-class, traversable structure rather than as metadata attached after retrieval: each evidence node is reachable through an explicit entity → filing → page → evidence chain, which is what makes the retrieval trace of Section 4.3 possible.

Listing 1 illustrates the intended representation for an FP&A variance context. The relation between a controller's narrative commentary and a reported variance is reified so that it can carry its own relevance type, confidence, validation status, and source provenance. The relevance type is what CARP consumes as a traversal weight: a CausalDriver edge propagates relevance at full weight, SupportingEvidence at reduced weight, and Correlational at a weight low enough that it is typically pruned by boundary detection beyond the first hop. The practical consequence is that an explanation assembled from this subgraph distinguishes the driver of a variance from facts that merely co-occur with it — a distinction flat retrieval cannot represent, because relevance there is a single undifferentiated similarity score.

**Listing 1. Contextual relevance and provenance annotation (design-level; see note below)**

```
:Q3_OPEX_Variance_BU3   a   kdaf:Variance ;
   kdaf:forPeriod          :FY2026Q3 ;
   kdaf:forBusinessUnit    :BU3 ;
   kdaf:varianceAmount     "1840000"^^xsd:decimal .

:Commentary_BU3_Q3_hiring   a   kdaf:NarrativeCommentary ;
   kdaf:text    "Contractor-to-FTE conversion ran ahead of plan in Q3." ;
   kdaf:author  :Controller_BU3 .

# Reified relation: typed, weighted, validated, provenance-bearing
:rel_0417   a   kdaf:ContextualRelation ;
   kdaf:from                     :Commentary_BU3_Q3_hiring ;
   kdaf:to                       :Q3_OPEX_Variance_BU3 ;
   kdaf:hasContextualRelevance   kdaf:CausalDriver ;   # | SupportingEvidence
                                                       # | Correlational
   kdaf:confidence              0.82 ;
   kdaf:validationStatus        kdaf:ExpertConfirmed ;
   kdaf:validatedBy             :Controller_BU3 ;
   prov:wasDerivedFrom          :Doc_BU3_Q3_CloseMemo_p4 .

# CARP traversal weights over relevance types (Section 2.6, Algorithm 1, w(r))
kdaf:CausalDriver        kdaf:propagationWeight 1.00 .
kdaf:SupportingEvidence  kdaf:propagationWeight 0.60 .
kdaf:Correlational       kdaf:propagationWeight 0.30 .
```

The representation is specified in OWL/RDF and materialized in a labelled property graph. The two levels serve different purposes. The RDF specification is the portable, standards-aligned artifact: it is what permits alignment with FIBO, reuse across financial domains, and inspection independent of any database vendor. Because RDF cannot annotate a triple directly, the relation in Listing 1 is reified as a first-class ContextualRelation resource. The property-graph materialization in Listing 2 expresses the same information natively as edge properties, which is what makes traversal-time weighting inexpensive: CARP reads the relevance type and confidence from the edge itself rather than resolving a reification node at every hop. The specification governs meaning; the materialization governs traversal.

**Listing 2. The same relation materialized as an annotated property-graph edge**

```
(c:NarrativeCommentary {id: 'Commentary_BU3_Q3_hiring'})
  -[r:EXPLAINS {
      relevanceType:     'CausalDriver',   // | SupportingEvidence | Correlational
      propagationWeight: 1.00,
      confidence:        0.82,
      validationStatus:  'ExpertConfirmed',
      validatedBy:       'Controller_BU3',
      derivedFrom:       'Doc_BU3_Q3_CloseMemo_p4'
  }]->
(v:Variance {id: 'Q3_OPEX_Variance_BU3', period: 'FY2026Q3', businessUnit: 'BU3'})
```

---

*Listings 1 and 2 are design-level. The FinanceBench graph evaluated in Section 4 instantiates the entity-anchoring and provenance subset of this representation (schema kdaf-financebench-provenance-v1); the contextual relevance types shown here are specified and in development but are not instantiated in the evaluated graph, because the benchmark contains no narrative commentary to which causal or supporting roles could be assigned. They are illustrated in the design scenario of Section 3 and are a target of the evaluation planned in Section 5.3.*

## 2.5 Stage 5: Hybrid Knowledge Validation

Knowledge accuracy is maintained through a hybrid validation pipeline combining automated and human-in-the-loop mechanisms. Automated consistency checking enforces ontology-level rules (e.g., EffectiveDate must follow PublicationDate). Relationships extracted with low confidence are flagged as Pending Validation. Domain experts validate flagged relationships through a targeted review interface, with decisions feeding back into model retraining. This cyclical validation ensures the knowledge base remains both scalable and trustworthy — crucial for enterprise FP&A contexts where decisions carry financial and regulatory implications.

## 2.6 Stage 6: Context-Aware Relevance Propagation (CARP)

To support LLM retrieval-augmented generation, the framework includes the Context-Aware Relevance Propagation (CARP) algorithm. Unlike standard graph traversals, CARP constrains expansion by ontology-defined relationship semantics and by hard analytical boundaries, and retains the decision record that produced its result. The process involves four steps: (1) Seed Node Identification — detecting entities relevant to a user query, each recorded with its match reason and confidence; (2) Weighted Propagation — expanding outward using ontology-defined relationship weights over relation types (in an FP&A deployment, the contextual relevance types of Section 2.4: CausalDriver, SupportingEvidence, Correlational; in the benchmark instantiation evaluated here, the entity-anchoring and provenance relation types described in Section 4.1); (3) Contextual Boundary Detection — limiting traversal based on a dynamic, depth-specific relevance threshold; and (4) Evidence Selection — ranking the reachable, eligible evidence and serializing a bounded number of items, with their provenance chains, as the contextual input for the LLM. In the profile evaluated here, step 4 returns an ordered set of evidence blocks rather than a connected subgraph; connectivity-aware assembly under an explicit context budget is the subject of the budgeted-assembly profile described in Section 5.4 and is not exercised in this paper. The traversal stage is serialized in full into a retrieval trace: which entities seeded the query and why, which edges were traversed and under which relationship type, which candidates were rejected and at what threshold and for what reason, and through which provenance chain each surviving evidence item is anchored to its source document. The final ranking stage is serialized for the items selected; evidence that is reachable and eligible but falls outside the evidence budget is counted but its individual non-selection is not currently recorded, a gap noted in Section 5.2. The trace is not a debugging artifact but the audit record evaluated in Section 4.2 and interpreted in Section 4.3.

Algorithm 1 states the procedure as implemented and evaluated. Three properties warrant comment. First, propagation is multiplicative and boundary detection is relative rather than absolute: the threshold at each depth is a fixed fraction of the best eligible score at that depth, subject to a floor, so the frontier adapts to how strongly the query anchors into the graph rather than to a schedule fixed in advance. Second, the company boundary is applied as a hard eligibility test during traversal, not as a post-hoc filter. Third — and this is the property that most distinguishes the implementation from an idealized graph retriever — selection is a two-stage process. Propagation determines which evidence is reachable and with what path score, but the evidence actually passed to the generator is chosen by an additive composite in which the normalized lexical score carries the largest single weight (0.45), with propagation score, concept coverage and period coverage contributing the remainder. CARP as evaluated is therefore a hybrid: graph structure governs eligibility, reachability and provenance, while final ordering is dominated by lexical evidence with ontology-derived signals reweighting within the reachable set.

Two consequences follow, both relevant to the results. Because lexical similarity dominates the final ordering, one should not expect large differences in answer content between this system and a strong lexical retriever, and Section 4.3 reports none. Because eligibility, reachability and provenance are governed entirely by the graph, one should expect differences in which sources are cited and in whether the selection can be explained, and Section 4.3 reports those. The design is deliberate — lexical matching is effective at locating candidate text and the ontology is what makes the selection accountable — but it means the evaluated system is not a pure graph retriever, and its results should not be read as though it were.

---

**Algorithm 1. Context-Aware Relevance Propagation (CARP)**

```
Input:  query q; graph G = (V,E); relation weights w; edge confidences c;
        company boundary company(q); max depth D; per-hop decay delta;
        threshold floor m; relative factor rho; frontier capacity B;
        evidence budget K
Output: ordered evidence context C; retrieval trace T

 1  q' <- ExpandQuery(q, company(q))
 2  l(e) <- normalised lexical score of evidence e against q'      // in [0,1]
 3  S_seed <- IdentifySeeds(q', G)    // company 1.0; ontology alias 1.0;
 4                                    // normalised period 0.9; each seed records
 5                                    // node type, score and match reason
 6  T.seeds <- S_seed ; best[s] <- score(s), path[s] <- [] for s in S_seed
 7  F <- sorted nodes(S_seed)
 8  for d = 1 to D do
 9     P <- {}
10     for each u in F, each incident edge (u, r, v) do
11         if v already occurs on path[u], skip            // no revisits
12         a <- NodeAffinity(v, q', l)
13         p <- best[u] * w(r) * c(r) * delta * a          // multiplicative
14         eligible <- not (v is evidence and company(v) != company(q))
15         P <- P + {(u, r, v, p, eligible)}
16     if P has no eligible proposal then break
17     tau_d <- max(m, rho * max{ p(x) : x in P, eligible(x) })
18     sort P by (-p, target_id, source_id, edge_type)      // deterministic
19     for each x in P do
20         reject x if not eligible(x), or p(x) < tau_d, or target already
21             accepted at this depth, or |A| = B, or p(x) <= best[target(x)]
22         else accept: best[target] <- p(x); path[target] <- path[u] + x
23         append (x, tau_d, decision, reason) to T.traversal_decisions
24     F <- sorted targets(A) ; if F empty then break
25  E_reach <- { evidence e reached : company(e) = company(q) }
26  for each e in E_reach do                               // final ranking
27     final(e) <- 0.45*l(e) + 0.20*best[e]
28              + 0.20*ConceptCoverage(e,q') + 0.15*PeriodCoverage(e,q')
29     provenance(e) <- canonical Company -> Filing -> Page -> Evidence chain
30  sort E_reach by (-final(e), citation_id)
31  C <- first K items of E_reach, serialised as evidence blocks
32  T.selected <- scores, best paths and provenance chains for C
33  return C, T
```

---

One scoping point governs how these results should be read. The evaluated condition instantiates Stages 3, 4 and 6 of the methodology — schema-guided extraction, contextual and provenance representation, and CARP retrieval — over a public benchmark. Stages 1, 2 and 5 (problem-centric scoping through competency questions, expert-validated bootstrapping of a minimum viable graph, and hybrid human-in-the-loop validation) presuppose domain experts and an organizational context that a public benchmark does not supply, and are not exercised here. The quantitative results that follow are evidence about the retrieval algorithm and its representation, not about the methodology as a whole.

Two implementations of KDAF are referred to in this paper and should not be conflated. The experimental implementation evaluated in Section 4 realizes Algorithm 1 in full and produces the retrieval traces reported there; it is the artifact against which every empirical claim in this paper should be checked. Separately, a reference implementation is maintained for practitioners applying the methodology in operational FP&A settings; it currently implements a deterministic governed profile of CARP — curated seed resolution over a validated minimum viable graph, with provenance and validation context attached — and does not yet perform weighted propagation or dynamic boundary detection. Section 5.4 describes the relationship between the two and the staged path by which the reference implementation adopts the algorithm evaluated here.

## 3. Design-Level Evaluation Scenario: Variance Analysis in FP&A

This section presents a design-level evaluation scenario demonstrating how KDAF can be instantiated in a realistic FP&A context. The purpose is to illustrate system behavior, scope, and interaction patterns. No quantitative performance claims are made here; empirical results are reported in Section 4.

### 3.1 Case Context

The scenario considers a synthetic but realistic FP&A environment modeled after a large U.S. enterprise operating multiple business units performing quarterly planning and variance analysis. Scope: 5 business units, approximately 50 cost centers, Q3 quarterly horizon, operating expense (OPEX) and contribution margin view. The analytical question posed by leadership: “Which cost drivers explain the variance versus plan in Q3?”

### 3.2 Knowledge Instantiation and Query Walkthrough

A subset of the enterprise financial ontology is instantiated focusing on entities relevant to variance analysis: BusinessUnit, CostCenter, ExpenseCategory, Forecast, Actual, Variance, and Period. Relationships capture allocation, aggregation, and attribution links between entities. Narrative commentary is associated with financial entities through contextual annotations indicating relevance to specific variances.

When the variance question is issued, CARP identifies relevant entities associated with Q3, Variance, and ExpenseCategory as seed nodes, then retrieves a focused subgraph covering: cost centers exhibiting the largest forecast-to-actual deviations; expense categories associated with those cost centers; business units aggregating the observed variances; and narrative annotations describing contributing operational factors. Entities not directly relevant to Q3 variance are excluded from the retrieved context.

### 3.3 Output Characteristics

The system produces a structured explanatory response grounded in the retrieved financial context: a concise narrative summary identifying primary cost drivers, with traceable references to underlying financial artifacts. Rather than presenting raw tables or aggregated totals, the output emphasizes explanatory structure, linking each identified driver to its associated business unit, cost center, and source documentation. Crucially, the response is accompanied by the retrieval trace that produced it, so a reviewer can verify not only what the system concluded but which evidence it consulted and why. This design-level walkthrough motivates the empirical evaluation in Section 4.

## 4. Empirical Evaluation

### 4.1 Experimental Setup

The proof-of-concept evaluation was conducted on FinanceBench [Islam et al., 2023], an open-source financial question-answering benchmark comprising analyst-grade questions over real company financial reports. The evaluation uses a frozen slice of 150 questions, of which 145 form the evaluation set (see below). The evaluation set is approximately balanced across the benchmark's own question-type field, with 49 novel-generated, 48 domain-relevant and 48 metrics-generated items. Five system conditions were compared:

- **llm_only**: No retrieved context — a direct measure of parametric LLM knowledge without retrieval.
- **bm25_text**: BM25 sparse retrieval [Robertson & Zaragoza, 2009] over the same evidence pool, no graph structure.
- **hybrid_text_table**: Lexical TF-IDF retrieval over the same pool with length normalization and a financial-concept term bonus, scoring text passages and table-derived evidence together. Despite the system name, this baseline uses no embedding model and no vector index; it is a concept-weighted lexical retriever, not a dense one.
- **graph_no_ontology**: Graph traversal over a knowledge graph constructed without ontological grounding.
- **graph_ontology / KDAF**: Ontology-grounded knowledge graph with CARP evidence retrieval (proposed).

All systems use Qwen3 [Yang et al., 2025] served locally via Ollama [Ollama, 2025] in a capped, no-think configuration, as the generative model. Two metrics were computed automatically over all 145 evaluation questions: correctness (exact-match or normalized numeric equivalence against gold answers; conservative by design) and traceability F1 (citation-level set overlap between predicted and gold source references). Binomial confidence intervals in Table 1 are Wilson score intervals [Wilson, 1927]; the paired system-versus-system differences in Table 2 use 10,000 deterministic paired bootstrap resamples over questions [Efron & Tibshirani, 1993]. Retrieval was executed once and cached; generation was then run three times (seeds 42, 43, 44) over the cached retrieval output. Because decoding is deterministic at temperature 0, the three passes produced byte-identical answers, so the reported figures carry no run-to-run variance — a reproducibility property, not a robustness result (Section 4.4). To calibrate scorer conservatism, a deterministic stratified LLM-assisted review was conducted on a representative 30-question sample per system using a Llama 3 adjudicator [Grattafiori et al., 2024] under a conservative policy (labeled llm_assisted_conservative).

A third group of measurements addresses auditability directly. The run serializes the full retrieval trace for every question: seed entities with match reasons and scores, traversal edges with relationship types, accepted and rejected candidates with depth-specific thresholds, and the provenance chain linking each selected evidence item to its company, filing, and page. From these traces we compute valid path rate (the share of selected evidence reachable by a well-formed typed path from a query seed), complete chain rate (the share whose provenance chain resolves end-to-end from company entity to source page), provenance resolution failures, retrieval replay fidelity across independent executions, cross-entity leakage, off-period evidence rate, and assembled-context token cost. The graph is constructed deterministically: two independent builds from the same inputs yield identical canonical digests, recorded in the environment report alongside the builder commit. Rejected-proposal counts differ by three orders of magnitude between the graph conditions (103,497 for KDAF against 38 for ungrounded traversal); this reflects search-space fan-out rather than selectivity of a different kind, since ontology concept and period seeds traverse reverse semantic edges across the shared evidence pool and 96,654 of KDAF's rejections are out-of-company candidates logged before rejection, whereas the ungrounded condition starts only from the subject company and follows a narrow provenance tree.

The source slice contained 150 questions. Five questions inspected during development of CARP's company-boundary behavior were designated calibration-only before the instrumented evaluation was run; the instrumented results and audit traces therefore use the remaining 145 untouched questions. No question was excluded because of execution or trace-serialization failure. The designation is recorded in the split manifest and experiment contract prior to execution, and all five system conditions produced complete outputs for all 145 questions across all three generation passes.

Traceability F1 is computed over sets of exact evidence identifiers. Predicted identifiers are those the generator emits in its answer, not the retrieval selections directly; the metric therefore scores whether the answer attributes itself to the correct sources, and a system may retrieve correct evidence yet score poorly by failing to cite it. Predicted citation strings and gold source strings are deduplicated into sets without normalization; a true positive is an exact string intersection. Precision is true positives divided by unique predicted evidence identifiers, recall is true positives divided by unique gold evidence identifiers, and F1 is their harmonic mean, defined as zero when precision and recall are both zero. Multi-source gold labels are handled setwise, so recall awards proportional credit for each matched identifier. The matching granularity is the evidence identifier, not the document or the page, and no normalization is applied — a deliberately strict choice that makes the metric a lower bound. The same strictness is what renders the metric inoperable on FinQA and TAT-QA under their current index granularity, as discussed in Section 5.2.

Two controls guard against the retrieval stage exploiting information it should not have. First, a leakage audit verifies that retrieval is invariant when question identifiers, gold answers, gold source labels, provenance metadata, and source-document metadata are removed from the index or replaced with adversarial values: if retrieval depended on any of these, its selections would change, and they do not. This forecloses the concern that a graph built from benchmark evidence might recover gold sources through label structure rather than through semantic matching. Second, retrieval is executed independently twice per question and the selections compared; agreement is complete across all systems and questions (Table 3). Both checks are emitted as machine-readable artifacts alongside the results.

All systems retrieve from a shared candidate pool of 189 evidence records assembled from the FinanceBench source index, applied identically across systems in identical order. The pool is gold-derived: retrieval never operates over full filings, but over pre-extracted evidence records comprising the gold evidence for the source slice together with the evidence of the calibration questions as distractors. Absolute values of traceability F1 and of the leakage rates reported below are therefore properties of this setting and are not comparable to open-corpus or full-document retrieval; the comparisons between systems, which share the pool exactly, are unaffected. Evidence associated with the five calibration questions remains in this pool: calibration questions are excluded from evaluation, but their evidence is retained as distractor context, since removing it would make the retrieval task easier than the setting the benchmark intends. A pool of this size relative to 145 questions constitutes a comparatively easy retrieval setting, a point returned to in Section 4.4.

Retrieval behavior differs across the graph conditions in one respect that bears on the results. In both graph systems, the subject company is enforced as a hard constraint: traversal cannot cross a company boundary, and the constraint is applied again before final selection. The reporting period, by contrast, enters graph_ontology as a soft scoring component rather than a hard filter, and is not used at all by graph_no_ontology. This asymmetry is a design choice with measurable consequences in both directions, quantified in Section 4.2.

## 4.2 Results

Table 1 reports headline results over the 145 evaluation questions; the graph_ontology (KDAF) row is highlighted. Table 2 reports paired differences, Table 3 the structural properties of the retrieval traces, and Table 4 retrieval-level behavior.

| System | Correct. | 95% CI | Trace. F1 | 95% CI | Reviewed | 95% CI | Latency (ms) |
|---|---|---|---|---|---|---|---|
| llm_only | 0.041 | [0.019, 0.087] | 0.000 | [0.000, 0.000] | 0.167 | [0.073, 0.336] | 1,216 |
| bm25_text | 0.117 | [0.075, 0.180] | 0.463 | [0.429, 0.495] | 0.233 | [0.118, 0.409] | 7,537 |
| hybrid_text_table | 0.117 | [0.075, 0.180] | 0.418 | [0.379, 0.457] | 0.300 | [0.167, 0.479] | 6,645 |
| graph_no_ontology | 0.103 | [0.064, 0.164] | 0.488 | [0.449, 0.524] | 0.233 | [0.118, 0.409] | 7,422 |
| **graph_ontology (KDAF)** | **0.110** | **[0.069, 0.172]** | **0.515** | **[0.480, 0.548]** | **0.300** | **[0.167, 0.479]** | **8,260** |

*Table 1. Evaluation results — FinanceBench, 145 questions per system. Correctness: exact-match or normalized numeric equivalence against gold answers (conservative by design; see Section 4.3). Traceability F1: mean per-question citation-level F1 over exact evidence identifiers, defined in Section 4.1. Reviewed correct: Llama-3-assisted conservative adjudication over a deterministic 30-question representative sample*

*per system; this calibrates the automatic scorer and is not independent human review. Intervals for correctness and reviewed correctness are Wilson score intervals computed from success counts; intervals for traceability F1 are 10,000-resample percentile bootstrap intervals over questions. llm_only produces no citations, so its traceability is identically zero. Latency is the mean of the three seed-wise median response times, of which retrieval accounts for under 5 ms (Table 4); unlike the accuracy measures it varies slightly across passes (KDAF range 8,124-8,478 ms). Correctness, traceability and reviewed correctness are identical across the three generation passes (seeds 42, 43, 44) because decoding is deterministic at temperature 0 — see Section 4.4.*

Table 2 reports paired system-versus-system differences. Because the same questions are answered by every system, paired resampling is the appropriate test: it removes between-question difficulty variance that dominates the unpaired intervals in Table 1.

| Comparison | Metric | Delta | 95% CI | Wins / losses / ties |
|---|---|---|---|---|
| KDAF - no ontology | Correctness | +0.007 | [-0.014, +0.034] | 2 / 1 / 142 |
| KDAF - no ontology | Traceability F1 | +0.027 | [+0.006, +0.050] | 11 / 3 / 131 |
| KDAF - no ontology | Reviewed correct | +0.067 | [0.000, +0.167] | 2 / 0 / 28 |
| KDAF - BM25 | Correctness | -0.007 | [-0.021, 0.000] | 0 / 1 / 144 |
| KDAF - BM25 | Traceability F1 | +0.052 | [+0.024, +0.083] | 23 / 4 / 118 |
| KDAF - BM25 | Reviewed correct | +0.067 | [-0.067, +0.200] | 3 / 1 / 26 |

*Table 2. Paired bootstrap differences, 10,000 deterministic paired resamples over questions. Correctness and reviewed-correctness comparisons are over 145 and 30 questions respectively. The two traceability intervals exclude zero; no correctness interval does. Wins, losses and ties count questions on which the first system scored higher, lower, or equal.*

Table 3 reports the structural properties of the retrieval traces for the two graph conditions. These are integrity checks rather than comparative findings: a correctly implemented deterministic retriever over a well-formed graph should attain perfect scores on all four, and the value of reporting them lies in their falsifiability — an earlier build did not attain them, as described in Section 5.5.

| System | Selected evidence | Valid path rate | Complete chain rate | Provenance failures | Mean path length | Replay fidelity |
|---|---|---|---|---|---|---|
| graph_no_ontology | 424 | 1.000 | 1.000 | 0 | 3.00 | 145 / 145 |
| **graph_ontology (KDAF)** | **426** | **1.000** | **1.000** | **0** | **1.28** | **145 / 145** |

*Table 3. Retrieval-trace structural properties over the 145 evaluation questions; graph schema kdaf-financebench-provenance-v1. Valid path rate: share of selected evidence for which a non-empty typed traversal path is recorded. The present audit verifies that a path exists and is serialized; it does not independently re-verify that the path originates at a recorded seed, is contiguous, and resolves every edge against the graph, so this figure should be read as a serialization completeness check rather than as an independent path validation. Complete chain rate: share whose provenance chain resolves company to filing to page to evidence. Provenance failures: selected items whose recorded provenance identifier does not resolve to that item (see Section 5.5). Replay fidelity: questions for which two independent retrieval executions produced identical selections, ordering and threshold decisions. Flat retrieval baselines emit ranked citation lists with similarity scores and cannot produce these quantities; they are not applicable rather than zero.*

Table 4 reports retrieval-level properties. These are computed from the retrieval stage alone and are invariant across generation seeds.

| System | Cross-entity leakage | Off-period evidence | Context tokens (mean / p50) | Retrieval p50 (ms) |
|---|---|---|---|---|
| llm_only | n/a | n/a | 0 / 0 | 0.0001 |
| bm25_text | 16.8% (73/435) | 23.0% (80/348) | 1,511 / 1,423 | 0.59 |
| hybrid_text_table | 20.2% (88/435) | 25.3% (88/348) | 1,402 / 1,254 | 0.75 |
| graph_no_ontology | 0.0% (0/424) | 28.2% (96/341) | 1,458 / 1,394 | 0.66 |

| System | Cross-entity leakage | Off-period evidence | Context tokens (mean / p50) | Retrieval p50 (ms) |
|---|---|---|---|---|
| **graph_ontology (KDAF)** | **0.0% (0/426)** | **20.5% (70/341)** | **1,581 / 1,551** | **4.14** |

*Table 4. Retrieval-level properties over the 145 evaluation questions; invariant across generation passes. Cross-entity leakage: share of selected evidence whose subject company differs from the question subject. Off-period evidence: among the 116 questions naming a fiscal period, share of selected evidence from a different period. Context tokens measured with the generator tokenizer (Qwen3-8B) and validated against the server prompt-token count. Retrieval p50 excludes generation. Graph systems enforce the company boundary as a hard constraint; in the lexical baselines company alignment is emergent rather than constrained.*

Two properties of the retrieval stage bear on the interpretation. First, the cost of graph retrieval is negligible relative to generation: CARP retrieval completes in under five milliseconds at the median, against generation latency measured in seconds, so the structure it adds is not paid for in user-visible latency. Second, the evaluated graph is small — 524 nodes and 1,852 edges, constructed in approximately 29 MB of resident memory — which is appropriate for the benchmark but is not evidence about behavior at enterprise scale.

Traversal path lengths differ systematically between the graph conditions. In graph_ontology, most selected evidence is reached at depth 1, because ontology-typed seeds — concept and period aliases resolved from the question — match evidence nodes directly; the ungrounded condition reaches essentially all of its evidence at depth 3, routing through the generic company-to-filing-to-page skeleton because it has no typed seed by which to enter the graph closer to the evidence. Shorter paths here indicate more direct grounding rather than truncation: the ontology supplies entry points that the unlabelled graph lacks.

Absolute accuracy on this benchmark is low for every condition, and should be read against published results rather than against intuition. Islam et al. evaluated sixteen model configurations on the same 150-case open-source sample and reported that GPT-4-Turbo used with a retrieval system incorrectly answered or refused to answer 81% of questions; they characterise the benchmark as a minimum performance standard that current systems do not meet [Islam et al., 2023]. Correctness in the 10-12% range under a conservative automatic scorer, rising to 23-30% under adjudication, is therefore consistent with the difficulty the benchmark was designed to expose rather than indicative of a malfunctioning system.

We stress that these figures are not directly comparable. The published results use substantially larger proprietary models, human review rather than automatic scoring or model-assisted adjudication, and retrieval over full filings; the present evaluation uses an 8B open-weight model in a capped configuration, a conservative exact-match scorer calibrated by a language-model adjudicator, and a small pre-extracted evidence pool that makes retrieval easier than the published setting. The comparison establishes that low absolute accuracy is expected on FinanceBench. It does not establish any ordering between this work and the published baselines, and no such ordering is claimed.

### 4.3 Interpretation

**First, retrieval is necessary.** Zero-context inference reaches 4.1% automatic and 16.7% reviewed correctness, well below every retrieval-augmented condition, and produces no citations at all. Parametric knowledge alone is insufficient for FinanceBench-style questions, and a system that cannot cite cannot be audited even in principle.

**Second — reported as a finding rather than deferred to limitations — graph structure does not pay for itself in answer correctness on this benchmark.** The four retrieval conditions span 0.103 to 0.117 automatic correctness with heavily overlapping intervals, and the paired comparison between KDAF and BM25 is -0.007 (95% CI [-0.021, 0.000]; 0 wins, 1 loss, 144 ties). On reviewed correctness KDAF and the lexical hybrid tie at 0.300 against 0.233 for BM25 and ungrounded traversal, but that interval is wide and does not settle the ordering either. The plain interpretation is that FinanceBench questions are predominantly single-document evidence-lookup tasks over a small candidate pool, a regime in which lexical matching is close to sufficient and relational structure has little opportunity to contribute. Claims that graph-

based retrieval generically improves answer accuracy over sparse retrieval are not supported here, and practitioners choosing an architecture for document-lookup workloads should weigh that.

**Third, on citation traceability the ordering reverses and the difference is statistically supported.** KDAF attains the highest traceability F1 of any condition (0.515), exceeding ungrounded graph traversal by +0.027 (95% CI [+0.006, +0.050]; 11 wins, 3 losses, 131 ties) and BM25 by +0.052 (95% CI [+0.024, +0.083]; 23 wins, 4 losses, 118 ties). Both intervals exclude zero. It is worth stating what a figure of this magnitude means operationally rather than leaving it as a comparative index. Traceability F1 is computed over exact evidence identifiers, so a score near 0.5 indicates that roughly half of the citation mass a reviewer would need to check aligns with the authoritative set: a controller verifying an answer would find a substantial share of the cited evidence correct and a comparable share either missing or extraneous. No condition measured here produces an answer whose citations can be accepted without inspection, and the paper does not claim otherwise; what the comparison establishes is that ontology-grounded retrieval reduces that verification burden relative to the alternatives, not that it eliminates it. Because matching is exact and unnormalized, the metric credits no near-miss — a citation to the correct page under a different identifier granularity scores as a miss — so these values are lower bounds on the alignment a human reviewer would recognise. This is the paper's central positive result, and its shape is instructive: the same systems that are indistinguishable on whether the answer is right are clearly distinguishable on whether the answer is attributable. Ontology-defined relationship weights act on which evidence is surfaced and how it is linked, not on the final answer string, which is precisely why the effect appears in traceability and not in correctness. The mechanism is visible in the ranking function itself (Section 2.6): lexical similarity carries the largest single weight in final evidence selection, so the answer content of KDAF and a strong lexical retriever converges, while the ontology-derived terms — propagation score, concept coverage, period coverage — together with hard entity eligibility reorder the selection within the lexically plausible set. The advantage this produces is precisely an attribution advantage, and it would be surprising if it were anything else.

**Fourth, graph-structured retrieval imposes an entity discipline that lexical scoring has no means of expressing.** Across the evaluation, KDAF selected no evidence belonging to a company other than the question subject — 0 of 426 items — while BM25 did so for 16.8% of its selections and the lexical hybrid for 20.2%. The company boundary in the graph conditions is an enforced constraint rather than a tendency; in the lexical baselines company alignment is emergent, and nothing in their scoring function could express the constraint even if it were wanted. The same pattern holds, less absolutely, for reporting period: KDAF returns off-period evidence for 20.5% of selections against 23.0% for BM25, 25.3% for the hybrid, and 28.2% for ungrounded traversal — the best rate of any condition, though far from zero, because period enters CARP as a soft scoring term rather than a hard filter (Section 4.1). Every selected item resolves to a complete provenance chain with zero resolution failures, and retrieval replays identically across independent executions for all 145 questions. Box 1 shows what this means for a single question; the rates above are what it means in aggregate.

**Fifth, the audit trail is not free, and one efficiency claim does not survive measurement.** KDAF assembles the largest retrieved context of any condition (1,581 tokens mean against 1,402 for the lexical hybrid) and has the highest median end-to-end latency (8.3 s against 6.6 s). A minimality reading of the retrieved context is therefore not supported: the assembled context is bounded and coherent, but it is not smaller than what lexical retrieval supplies, and we make no efficiency claim for it. Compactness under an explicit context budget is a property of the budgeted-assembly profile described in Section 5.4, which this evaluation does not exercise. Two components of the cost are worth separating, however. Retrieval itself is negligible — 4.1 ms at the median, under 0.1% of response time — so the additional latency is attributable to generating over a larger context, not to graph traversal. Whether the additional tokens are well spent is answered by the traceability and discipline results above, not by the token count.

**Finally, the automatic scorer substantially underestimates correctness.** Reviewed correctness runs roughly two to three times higher than automatic correctness across retrieval conditions (KDAF: 0.110 to 0.300), reflecting exact-match scoring's inability to credit semantically equivalent paraphrases and partial multi-part answers. Automatic figures are lower bounds. The gap is not perfectly uniform: BM25 leads on automatic correctness and trails on reviewed correctness, while KDAF does the reverse. Given the width of the 30-question reviewed intervals neither ordering is established, which is itself the point — under either instrument the correctness comparison fails to discriminate between these systems. The traceability metric is conservative in the same direction: it requires exact evidence-identifier equality with no normalization, so it credits no partial or coarser-grained match.

---

### Box 1. What an auditable retrieval trace looks like

**Question** (FinanceBench, financebench_id_08286): *"By drawing conclusions from the information stated only in the income statement, what is Amazon's FY2019 net income attributable to shareholders (in USD millions)?" Both systems answer correctly ($11,588 million). The difference is what each can show an auditor.*

**BM25 sparse retrieval — ranked citation list (top 3 selected):**

- NIKE_2019_10K:p53 — score 31.59, **selected** ← a different company's filing, top-ranked on lexical similarity
- AMAZON_2019_10K:p37 — score 31.58, selected (gold source)
- AMERICANWATERWORKS_2021_10K:p85 — score 30.16, selected

*Audit trail ends at similarity scores. Nothing in the trace explains — or prevents — the consultation of Nike's and American Water Works' filings for an Amazon question.*

**KDAF (graph_ontology) — CARP retrieval trace:**

- **Seed nodes (with match reasons):** company:amazon (Company; question_company_exact_match, 1.000); concept:net_income (FinancialConcept; ontology_alias_match, 1.000); period:fy2019 (Period; normalized_period_match, 0.900)
- **Top evidence (final score 0.980; propagation score 0.902):** Traversal: concept:net_income —MENTIONS_CONCEPT→ AMAZON_2019_10K:p37 (gold source). Provenance: company:amazon —HAS_FILING→ amazon_2019_10k —HAS_PAGE→ p.37 —CONTAINS_EVIDENCE→ evidence.
- **Boundary decisions:** 16 traversal proposals accepted, 441 rejected; depth thresholds {1: 0.406, 2: 0.373, 3: 0.281}, each computed as 0.45 of the best eligible score at that depth subject to a 0.08 floor. Scores quoted above are final composite scores; propagation scores are given alongside for comparison.
- **Exposed design consequence:** items 2 and 3 are Amazon FY2017 evidence (final score 0.748; propagation 0.839) selected despite the FY2019 period seed. This follows directly from the constraint asymmetry noted in Section 4.1: company is enforced as a hard boundary, period is only a soft scoring term. The trace makes the consequence visible and attributable to a specific design decision, and Table 4 quantifies its aggregate cost.

*Every consulted document is reached through a typed path from the question's entities. The company boundary is enforced, not merely typical: across the 145 evaluation questions KDAF selected 0 of 426 evidence items from a company other than the question subject, while BM25 did so for 73 of 435 selections (16.8%) and the lexical hybrid for 88 of 435 (20.2%). The point is not that flat retrieval is careless. A production sparse stack can of course pre-filter candidates on a company field; what it cannot do is derive that constraint from the same entity model that supplies the audit trail. In the graph conditions the boundary and the provenance chain are two consequences of one representation, rather than a filter bolted onto a scoring function with no notion of entities at all.*

*Traces are drawn from the evaluation run itself, over the 145 untouched questions (see Section 4.1). The FinanceBench graph instantiates the entity-anchoring and provenance subset of the KDAF ontology (schema kdaf-financebench-provenance-v1). The contextual relevance types of Section 2.4 and Listing 1 (causal, supporting, correlational) are specified and in development but are not instantiated in this graph;*

*the auditability demonstrated here therefore rests on entity anchoring and provenance completeness alone, and is a lower bound on what the full representation is designed to expose.*

---

### 4.4 Limitations of the Pilot

Three further limitations concern the evaluation setting itself. The baseline set contains no dense-vector retriever: both non-graph baselines are lexical, and the concept-weighted hybrid condition despite its name uses no embedding model. Comparisons against embedding-based retrieval, and against published graph-RAG systems, are therefore outside what this evaluation establishes. The candidate pool is small — 189 evidence records shared across 145 questions — which makes retrieval comparatively easy and plausibly favors lexical matching; results should not be extrapolated to settings with large or noisy candidate pools. Finally, the constructed graph contains 524 nodes and 1,852 edges, appropriate to the benchmark but far below enterprise scale, so no claim about scaling behavior is supported by these measurements.

Five further limitations govern the scope of claims. First, the evaluation is conducted on a single benchmark (FinanceBench); results on FinQA and TAT-QA remain diagnostic (see Section 5.2). Second, at 145 questions the correctness comparison is underpowered by construction rather than by sample size alone: 142 of 145 paired correctness comparisons between the graph conditions are ties, and 144 of 145 between KDAF and BM25, so the discordant-pair rate bounds what any same-benchmark scaling could resolve. This is a property of the benchmark rather than of the method, and it is why Section 5.3 proposes a different evaluation instrument rather than a larger slice of this one. Third, automatic correctness is systematically conservative, with reviewed scores two to three times higher, so Table 1 correctness figures are lower bounds. Fourth, all experiments use a single generative model (Qwen3-8B via Ollama in a capped, no-think configuration) and a single adjudicator (Llama 3). Generalization across model families is untested, and two consequences of these choices bear on the results: a capped, no-think configuration constrains reasoning depth and may compress differences between context qualities that a stronger generator would exploit, and the reviewed-correctness adjudicator is itself an unvalidated language model, with no human agreement study performed against its labels. Reviewed correctness should therefore be read as a calibration of the automatic scorer by a second, independently conservative instrument, not as ground truth. Fifth, decoding is deterministic at temperature 0, and the three generation passes reported here produced byte-identical answers. This establishes end-to-end reproducibility but is explicitly not evidence of robustness under stochastic decoding, which remains untested; readers should not interpret the absence of seed variance as stability under sampling.

## 5. Discussion

### 5.1 Scientific Implications

The proposed KDAF framework contributes to several intersecting research areas. Within knowledge graph engineering, the minimum viable graph approach offers a concrete, business-question-driven operationalization of the cold-start problem that complements existing ontology construction methodologies [Meckler, 2024; Kommineni et al., 2024]. Unlike top-down ontology design, which begins with comprehensive schema development, the problem-centric scoping stage in KDAF bootstraps a graph from a minimal, validated scaffold — a pattern that may generalize to other high-stakes analytical domains requiring expert-in-the-loop knowledge modeling.

Within retrieval-augmented generation, this work makes a claim that is architectural rather than metric-driven. The evidence presented here does not support the proposition that graph-structured retrieval is generically more accurate than sparse retrieval; on single-document filings questions it is not. What it does support is that grounded retrieval and flat retrieval differ in kind with respect to what they can be held accountable for. A similarity score is a ranking justification, not an explanation; a typed, provenance-complete traversal path is an audit record. In domains where a wrong answer is a compliance event rather than an inconvenience, the ability to reconstruct why evidence was consulted is not a secondary property to be traded against a small accuracy delta — it is a precondition for deployment. This reframing has a methodological consequence: the field's dominant evaluation instruments (exact-match correctness, citation-overlap F1) are blind to it, and measuring auditability requires metrics that do not yet exist in standard benchmark suites.

More broadly, the framework illustrates how symbolic and statistical AI can be composed in a complementary rather than competitive manner. While developed in the FP&A context — where auditability and traceability carry regulatory and organizational weight — the underlying methodology may transfer to other enterprise domains requiring high-stakes analytical reasoning over heterogeneous, semi-structured data, including supply chain analytics, healthcare operations, and regulatory reporting.

## 5.2 Limitations

Several limitations govern the scope of claims in this work. First, the framework requires an initial ontology bootstrapping effort involving collaboration between domain experts and technical practitioners. Although the minimum viable graph approach mitigates the cold-start problem, this upfront investment may represent a barrier for organizations with fragmented financial definitions or limited semantic modeling experience, and this paper does not quantify that cost.

Second, the effectiveness of the approach is contingent on underlying data quality and governance. The framework improves transparency and traceability but does not correct inaccuracies at the source. Poorly governed financial data or inconsistent planning assumptions can propagate through the knowledge graph, limiting the reliability of generated explanations.

Third, the strength of the evidence differs by claim and should not be summarized in a single phrase. The correctness comparisons are underpowered by the benchmark's tie structure and are directional at best. The traceability differences carry 95% confidence intervals excluding zero under paired resampling and are statistically supported within this setting. The retrieval-discipline and provenance results are exhaustive counts over the full evaluation rather than estimates. What these share is narrowness of setting, not weakness of measurement.

Fourth, research question RQ1 asks whether the methodology reliably guides knowledge graph construction. This paper addresses RQ1 at the level of design and argument only: as noted in Section 4.1, the stages that would have to be evaluated to answer it — competency-question scoping, expert-validated bootstrapping, and human validation loops — are not exercised by a public benchmark. A construction study with domain experts, measuring competency-question coverage, extraction precision against gold annotation, and expert validation workload, is required to answer RQ1 empirically and is planned as part of the work described in Section 5.3.

Fifth, the framework assumes the presence of human validation loops, particularly during early deployment. Automated extraction and reasoning are complemented by expert review to resolve ambiguous relationships and low-confidence assertions. Full automation is neither expected nor appropriate in regulated financial environments.

Sixth, the structural auditability property emphasized in Section 4.3 is demonstrated but not yet valued: citation-overlap F1 rewards retrieving the right documents, not exposing an inspectable reasoning path, and no metric in this evaluation measures whether an auditable trace shortens human verification, improves reviewer confidence, or catches errors that would otherwise pass. Metrics for path faithfulness, explanation consistency under paraphrase, and human verification effort are required before the practical value of that advantage can be claimed.

Seventh, two of the three planned external benchmarks — FinQA [Chen et al., 2021] and TAT-QA [Zhu et al., 2021] — are currently at diagnostic rather than headline status. For FinQA, evaluation reveals both genuine retrieval failure (the non-oracle candidate pool uses passage-level chunks that do not align with the table-row support units required for multi-step numeric program execution) and an incompatible source-ID granularity between the retrieval index and gold labels; correcting these requires a structured table-to-support mapping and index reconstruction before traceability claims can be made. For TAT-QA, a source-ID prefix mismatch between the non-oracle pool and gold labels depresses traceability to zero in the scorer even when retrieved context is substantively correct; prefix normalization would recover approximately 12–13 traceability matches per 100 questions, but both scorer repair and retrieval-index improvements are needed before these results enter the main table. TAT-QA automatic correctness (14–17% for retrieval-bearing systems) is valid under the current answer scorer and is available as supporting evidence; full headline status for both benchmarks is deferred to the subsequent extended manuscript.

Two limitations concern the audit machinery itself rather than the results. First, trace completeness is asymmetric across the two retrieval stages: every traversal proposal is recorded with its threshold and rejection reason, but at the final ranking stage only the selected items are serialized, so evidence that was reachable and eligible yet fell outside the evidence budget is counted without an individually recorded reason for its exclusion. A trace intended as an audit record should record both, and extending serialization to the ranking stage is a required next step. Second, the valid-path figure in Table 3 verifies that a typed traversal path was recorded for each selected item; it does not independently re-verify that the path begins at a recorded seed, is contiguous, and resolves every edge against the graph. It is therefore a serialization check rather than an independent validation, and a stronger audit should perform the reconstruction rather than trusting the record.

Finally, two limitations follow from the choice of benchmark. The framework is motivated by FP&A and evaluated exclusively on public filings question answering; no FP&A planning data, variance structure, or narrative commentary was evaluated, and the scenario in Section 3 is a design illustration rather than an executed system. Claims about FP&A behavior therefore rest on argument, not measurement. Relatedly, the off-period evidence rate of roughly one in five selections, while the lowest of any condition measured, would be difficult to defend in an operational financial reporting context, where evidence drawn from the wrong fiscal period is a material error rather than a ranking imperfection. Promoting the reporting period from a soft scoring term to a hard eligibility constraint, as the company boundary already is, is the most direct remedy and is the first change we intend to evaluate.

### 5.3 Future Work

The primary next step follows directly from the repositioning above: if auditability is the axis on which ontology-grounded retrieval earns its cost, then auditability must be measured rather than asserted. Four lines of work follow.

First, metric development. We plan to operationalize path faithfulness (the share of answer statements attributable to the retrieved evidence and its traversal paths), explanation consistency (stability of answers and citation paths under question paraphrase), and reviewer verification effort (time and error rate for a finance professional validating an answer with versus without a retrieval trace). The last of these requires a practitioner study with FP&A analysts and is the most direct test of whether structural auditability delivers practical value.

Second, benchmark construction. FinanceBench rewards single-document lookup and therefore cannot exercise the relational semantics KDAF is designed for. We plan a synthetic-but-realistic FP&A variance-attribution benchmark with known causal ground truth — a ledger of business units, cost centers, and accounts with plan-versus-actual by period, injected variance drivers (price, volume, FX, one-off, reclassification), narrative commentary linked to drivers, and a subset containing deliberately conflicting commentary. Question types would span variance attribution, multi-period trend analysis, allocation tracing, and conflict resolution, making the benchmark itself a releasable contribution.

Third, strengthening the audit machinery. Serialization will be extended to the final ranking stage so that non-selected reachable evidence carries a recorded reason, and the path audit will be reimplemented to reconstruct each traversal from the graph rather than accepting the recorded path, closing the two gaps identified in Section 5.2.

Fourth, algorithmic and empirical strengthening. Planned work includes internal CARP ablations isolating the contribution of relationship-type weights, dynamic thresholding, and seed identification; a sensitivity sweep over relevance weight parameterization; the addition of dense-vector and published GraphRAG baselines, together with a sparse baseline restricted to the question subject entity by metadata filter — the direct counterfactual for the retrieval-discipline result of Section 4.3; and replication across multiple model families and random seeds. Results from these evaluations will be incorporated into an extended manuscript targeted at peer-reviewed venues in enterprise AI, semantic computing, or applied knowledge graph systems.

### 5.4 Reference Implementation and Adoption Path

A methodology intended for enterprise practice should be assessable by practitioners, not only by reviewers. Alongside the experimental implementation used for the evaluation in Section 4, an open reference implementation of KDAF is maintained for FP&A teams applying the six-stage methodology to their own data. The two are deliberately distinct: the experimental implementation optimizes for controlled measurement and complete decision traces, whereas the reference implementation optimizes for governed operation — read-only access to the system of record, validated competency questions, citation checking, and explicit refusal when a generated claim is not supported by retrieved evidence.

The reference implementation exposes CARP in three staged profiles. The deterministic governed profile, available now, resolves stored competency questions to curated concepts in a validated minimum viable graph and assembles evidence with provenance and validation state; it is high-precision and fully reproducible, but performs no scoring or propagation. The generalized propagation profile corresponds to Algorithm 1 as evaluated in this paper: intent-conditioned relationship weighting, best-first multi-hop expansion, dynamic depth-specific boundaries, and a complete accept-and-reject decision trace. The budgeted assembly profile extends selection from ranked evidence to a connected subgraph chosen under a serialization budget, maximizing concept coverage and provenance completeness while penalizing redundancy. Staging the profiles in this order allows each addition to be evaluated against the deterministic baseline rather than assessed only by inspection of individual answers.

This separation also bounds what the present evaluation demonstrates. The results in Section 4 establish properties of the algorithm under controlled measurement; they are not evidence about the reference implementation's current retrieval behavior, which implements an earlier profile by design.

### 5.5 Code and Data Availability

The experimental implementation constructs its knowledge graph in memory from the released inputs and requires no external graph database. Reproduction therefore needs only the interpreter, the graph build script, the configuration, and the evaluation split — there is no database server to provision, schema to migrate, or vendor dependency to satisfy. The graph snapshot digest and build-script version are recorded in the environment report so that the graph itself, and not merely its description, can be rebuilt.

One methodological note about these artifacts is worth recording, because it is easy to under-value negative checks. The provenance audit was specified to report resolution failures explicitly rather than to report only a success rate. On first execution it returned a non-zero failure count, revealing that shared graph edges stored a single citation identifier that later evidence records could overwrite: chains resolved topologically while the recorded identifier could belong to a different evidence item. A success-rate-only check would have reported complete resolution and the defect would have reached publication inside the paper's central claim. The representation was corrected before the run reported here, which records zero resolution failures across all selected evidence (Table 3). Audits that can only confirm are not audits.

The evaluation is supported by a set of audit artifacts produced as part of the experimental protocol rather than after the fact. These include the pre-registered split manifest and experiment contract recording the calibration and evaluation partitions before execution; the run-level trace audit, which is the source for the structural figures of Table 3; the environment report capturing model versions, configuration, and run identifiers; the retrieval-trace audit; the package audit verifying row parity and artifact completeness; and the retrieval leakage audit described in Section 4.1. Configuration files, prompts, the ontology schema, the evaluation split, the environment report and the audit artifacts described above are archived as a citable deposit (Zenodo, DOI 10.5281/zenodo.22022068), which also includes the graph builder, retrieval, scoring and audit scripts required to reproduce the reported run. Benchmark-derived question text, gold labels, and evidence passages are not redistributed: the source benchmark publishes no explicit licence grant, so those materials remain subject to the originating dataset's terms and are referenced by identifier only. The deposit includes a script that reconstructs the evaluation inputs from the upstream benchmark repository and verifies them against the recorded evaluation-split and graph digests, so a reader can rebuild the exact inputs used here without relying on redistributed content.